# Conditional Generative Adversarial Networks for Emoji Synthesis with Word Embedding Manipulation

**N. Dianna Radpour** and **Vivek Bheda**
Department of Linguistics, Department of Computer Science
State University of New York at Buffalo
{diannara, vivekkan}@buffalo.edu

**Abstract**

Emojis have become a very popular part of daily digital communication. Their appeal comes largely in part due to their ability to capture and elicit emotions in a more subtle and nuanced way than just plain text is able to. In line with recent advances in the field of deep learning, there are far reaching implications and applications that generative adversarial networks (GANs) can have for image generation. In this paper, we present a novel application of deep convolutional GANs (DC-GANs) with an optimized training procedure. We show that via incorporation of word embeddings conditioned on Google's word2vec model into the network, the generator is able to synthesize highly realistic emojis that are virtually identical to the real ones.

## 1 Introduction

Generative adversarial networks have seen lots of recent success in image generation. By training two separate networks, a generator and discriminator, simultaneously, samples are generated to then be continuously scrutinized and improved, until the discriminator can no longer discriminate the generated samples as being synthesized [3]. The potential of GANs has sparked a lot of deep learning research in the realm of image generation, since the framework can be applied to different realms with ease.

Another area that has seen a surge of interest and overwhelming popularity in digital communication and the age of online personas is the presence of emojis. Emojis are appealing to use when communicating since they can easily be incorporated into any message, post, or comment, and are also free from any language barriers that would pose problems for non-pictorial communication. They are also widely used among different social media platforms and recently companies who have began making their own brand emojis. As such, they are constantly being updated to include more variety and to have broader inclusivity. However, an inevitable shortcoming of emojis is their inability to have a truly complete and comprehensive representation. As of recent, there were a total of 722 emoji characters available in the standard Unicode character set, but only 82 were emotive face ones. To think that the entire human experience of emotion can be captured into 82 singular images is a hefty assertion, and the task of continuously updating and adding face emojis one by one is laborious and inefficient. With the emergence of GANs, it seems a plausible task to automate the process of emoji generation, and implement them with word embeddings that are associated with the images of the desired target emojis to create highly realistic emojis.

## 2 Related Work

Since their advent in 2014 by Goodfellow et. al, GANs have proven to be quite fruitful in image generation, and have been successfully implemented for a range of different synthesis tasks in just the last couple years. The networks are made up of the generative model that is pitted against an adversary discriminative model that learns to distinguish whether a sample is from the model distribution (an artificially generated one) or from the data distribution (a real one). The generative model generates samples via passing random noise through a multilayer perceptron, and then sending those samples to the discriminative model which is also a multilayer perceptron. [3]

Recently, Reed et. al [10] showed that GANs could be used to generate images from text. They were able to generate extremely real looking flowers and birds based on written descriptions, by using deep convolutional and recurrent text encoders that learned a correspondence function with images. Their results with this approach were quite impressive, and showed how the power of detailed text descriptions could be harnessed for generative adversarial training to generate highly compelling images.

There have also been more and more variations seen in the implementation of GANs. Zhang et. al [12] presented a Stacked Generative Adversarial Networks (StackGAN) with the goal of generating high-resolution photorealistic images. In their model, the Stage-I GAN would sketch the primitive shape and colors of the object based on given text description, which would yield low-resolution images. However, then the Stage-II GAN of their model would take the outputs of Stage-I along with their text descriptions as inputs, and would generate high-resolution images with photo-realistic details.Their sketch-refinement process showed to be quite successful and produced very realistic images.

Bao et. al [2] have recently presented variational generative adversarial networks, which is basically a general learning framework that combines a variational auto-encoder with a generative adversarial network. Their network works by synthesizing images in fine-grained categories, such as faces of a specific person or different objects, and demonstrates how a variational approach to GANs can be used in generating realistic and diverse samples that have fine-grained category labels.

Antipov et. al [1] have recently shown that GANs can be implemented to produce images for automatic face aging via the "Identity-Preserving" optimization of GAN's latent vectors to incorporate face recognition and age estimation solutions. Their results were astounding, and the incredibly realistic aged and rejuvenated faces demonstrated the high potential of the proposed method.

Wang et. al [11] have recently presented a new method for synthesizing high-resolution photo-realistic images from semantic label maps using cGANs, which high-resolution results that proved promising for the realm of deep image synthesis and editing. Osokin et. al [7] have also recently proposed a novel biomedical application of GANs for the synthesis of cells imaged by fluorescence microscopy that had impressive results that are sure to have great implications for biomedical engineering in the years to come.

## 3  Methods

Our general approach was to train a deep convolutional generative adversarial network (DC-GAN) conditioned on emojis with their associated text embeddings, in the form of word vectors from Google's word2vec model. It was then fed into a convolutional neural network in the generator network G, followed by the discriminator network D performing feed-forward inference conditioned on all the word embeddings.

### 3.1  Generative Adversarial Networks

The general architecture of GANs consists of multiple convolutional and dense layers. The training procedure in GANs is similar to a two-player min-max game with the following objective function,

$$\min_G \max_D V(D, G) = \mathbb{E}_{x \sim p_{data}}[\log D(x)] + \mathbb{E}_{z \sim p_z}[\log(1 - D(G(z)))],$$

The discriminator in our network was conditioning our model on the text using three elements: real images and true labels, real images and fake labels, and fake images and the real labels. The loss from this process then gets propagated to the generator so the gradient is upgraded based on the average loss from those. Via this introduction of text embeddings, we follow Reed et. al's [8] example to optimize the following structured loss:

$$\frac{1}{N} \sum_{n=1}^{N} \Delta(y_n, f_v(v_n)) + \Delta(y_n, f_t(t_n))$$

### 3.2  Deep Convolutional GANs

While CNNs with unsupervised learning were initially approached with reluctance, Radford et. al [8] presented a new class of GANs that helped to bridge the gap between the success of CNNs for supervised learning and unsupervised learning. They introduced DC-GANs, which have certain architectural constraints, and have shown to be quite successful for unsupervised learning. In the DCGAN, the generator takes a randomly generated noise vector as input data and then uses a technique called deconvolution to transform the data into an image. The discriminator is a classical convolutional neural network, which classifies real and fake images. In our generator we had one vector that was random noise and one with the word embeddings. The word embedding input vector is connected to the fully-connected layer to reduce its size. DCGANs produce better images because they don't use any pooling layers or any fully-connected layers while producing the images. The deep convolutional adversarial pair is able to learn a hierarchy of representations from the images and word vectors in both the generator and discriminator.

### 3.3  Word Embeddings

Our network was created via generative adversarial training conditioned on word vectors from Google's word2vec model. This was done by assigning words to each of the emojis that, via human intuition, were thought to be most representative of the emoji at large. Then with word2vec, each of the emojis got a respective association with a word vector.

## 4  Dataset

Our dataset consisted of 82 commonly-used face emojis, used in Facebook Messenger. These emojis appear for users of Facebook, while iOS and Android users have native emojis for their respective platforms on the app.

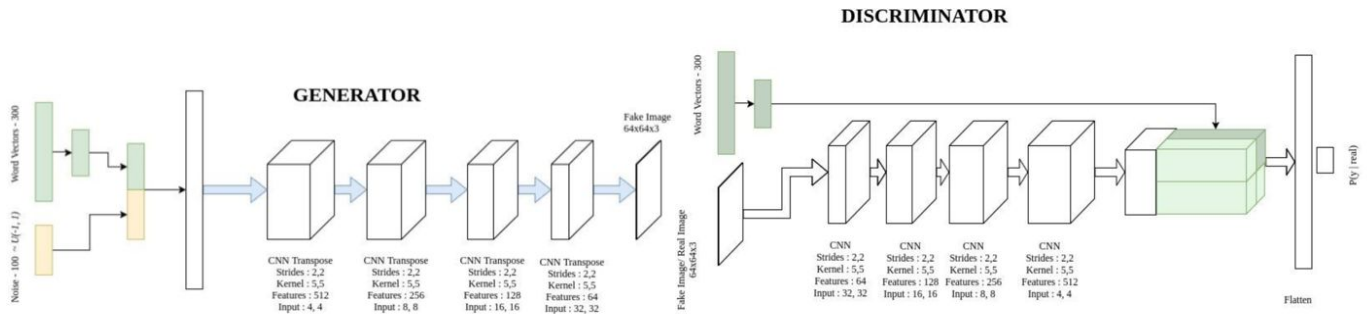

Figure 1. The general model architecture - a generator connected to a discriminator (as in most DC-GANS) but we changed convolutional layers

## 5 Model Architecture

Our model architecture is consistent with that originally proposed by Radford et. al for DC-GANs [8]. The three main things that these networks have that contributes to their unique potential and applications is the all convolutional net that replaces deterministic spatial pooling functions, an attempt at elimination of the fully connected layers on top of convolutional features, and batch normalization to stabilize learning by normalizing the inputs. Our architecture includes these aspects of the deeper GAN models, as demonstrated in Figure 1. The generator takes noise in the form of a randomly generated noise vector for input data along with the real labels, and then uses a technique called deconvolution to transform the data into an image. That output image is then fed to the discriminator, which is a classical convolutional neural network that aims to classify the real and fake images. The loss from this process then gets propagated to the generator for the update and the gradient is upgraded based on the average loss from those.

### 5.1 Training

The training procedure for the discriminator in our network began with conditioning our model on the text using three elements: real images and true labels, real images and fake labels, and fake images and the real labels. Sometimes it would be necessary to train the generator twice because the discriminator would train to point of optimum too quickly and cause the gradients to vanish. After training, we looked at what the generator's outputs for the different inputs (different word vectors) were. Overall, the training procedure for the discriminator network was split into three parts: real images with real labels, real images with fake labels, and fake images with real labels, and the weighted loss of the three combined became the loss that was used to update the discriminator. In graphs (a) and (b), the loss of both the generator and discriminator during training is shown.

(a)

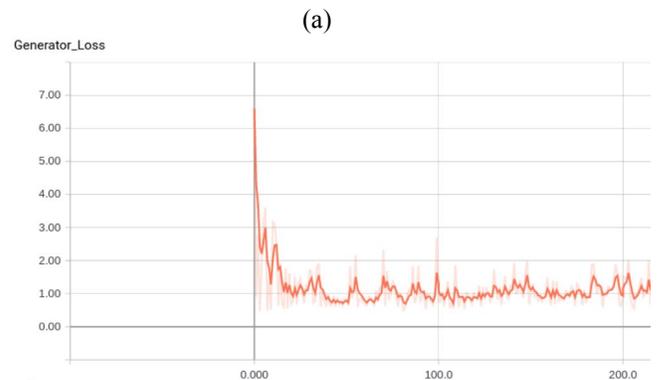

(a) The loss of cGAN generator during training

(b)

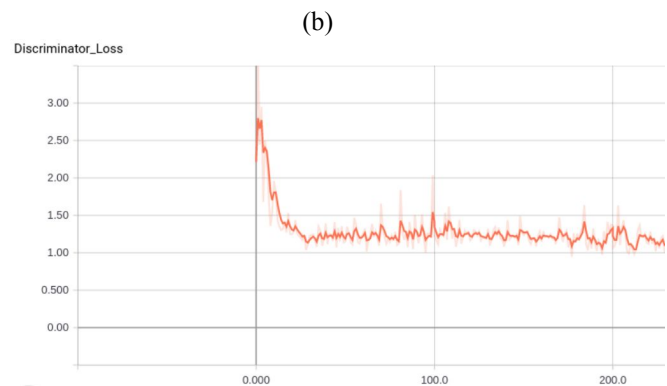

(b) Loss of cGAN discriminator during training

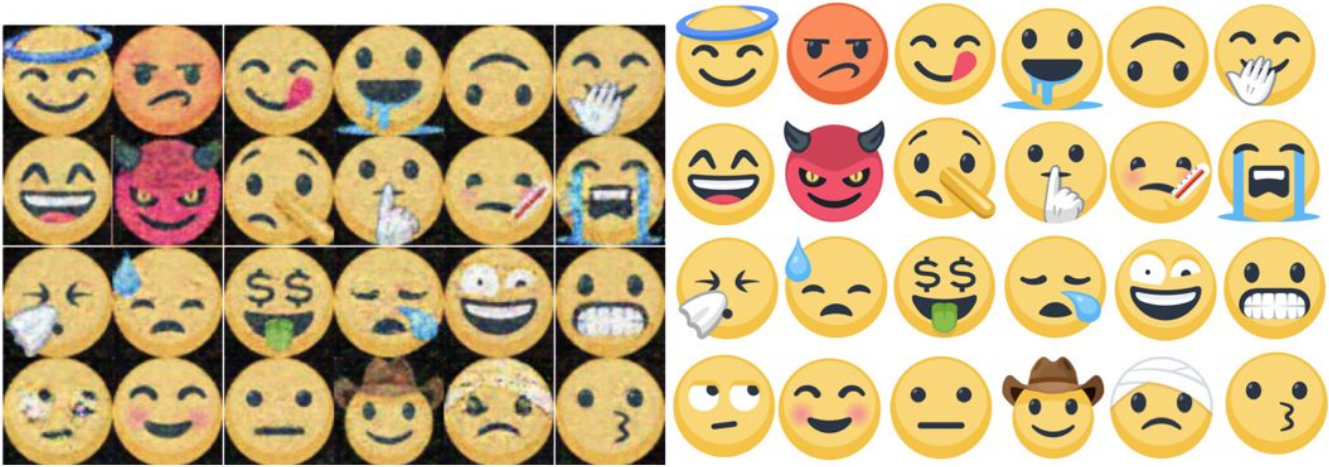

Figure 2. Examples of emojis generated with conditioning the network on associated word embeddings (on left) with the respective real emojis (on right)

### 5.2 Early Stopping

Our network was regularized with early stopping, where we essentially just stopped the training procedure when we noticed the quality of images becoming good and strongly correlating with the labels. In the DC-GAN, we often found that training too long basically caused it to reset and learn a new set of features from scratch. Because of this, every 1000 or so epochs, the quality of the images being generated would become equivalent to noise, and so updating the learner was necessary for better fitting.

## 6 Experiments and Results

### 6.1 Implementation

Initially, we tried arithmetic in the word vector space by averaging the word vectors of two different emojis to test if we could synthesize with features of both. Features from both emojis did end up in the result, but not in a realistic manner. The generator would output images that had features from both the original emojis of the original word vectors, but these were quite noisy, though you can make out the features from the two original emojis in Figure 3.

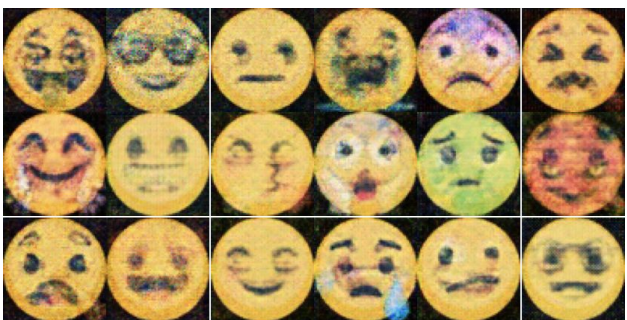

Figure 3. Emojis generated without conditioning the network on associated word embeddings.

### 6.2 Evaluation

Initially, we observed a lot of noise in the samples generated without conditioning on the individual word embeddings. However, as Figure 2 illustrates, the emojis that were generated by the DC-GAN with word vectors that we manually assigned to each of the emojis showed very realistic results. While a few of the emojis generated by our network are a bit grainy and have a few blurred features, for the most part they are incredibly similar to the real emojis.

## 7 Conclusion

In this paper, we described an approach that optimized the procedure by which generative adversarial training can be implemented for image generation by conditioning the network on word vectors from Google's word2vec model. We can see from our results that the vector space is clearly mapping to the features of the emojis. We found that conditioning the GAN yielded much better results and is evident that the included word embeddings played a huge role in synthesizing the realistic samples.

## 8 Future Work

Seeing the success that DC-GANs had in synthesizing such realistic face emojis with almost negligible differences when compared side-by-side to the real ones, speaks to the potential of this approach and we hope to extend the task to synthesizing new emojis with compound emotions. With the incorporation of Reed et. al's [9] new model, the Generative Adversarial What-Where Network (GAWWN), that synthesizes images given instructions describing what content to draw in which location, it seems very possible to be able to tackle the same task that was attempted in Figure 3, but restricting the location of certain features of the emojis. For example, with inputs being an unseen phrase, like 'food coma' that have no emoji associated with them, being able mix the the features of the following two emojis, and synthesize an emoji with the mouth features of the first one, and the eye features of the second one (simply via instructions regarding the location of the face that the respective content should go to), and hopefully being able to synthesize something as follows:

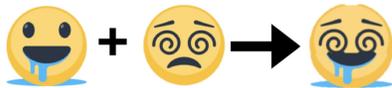

Controlled adversarial image generation like so has a lot of potential in creating new and unseen images, and the potential applications are truly encouraging.

Additionally, we hope to incorporate emojis from different sources, such as Google's emoji set, into our network so that the random noise would start mapping to the style of the emoji and the vector space would map to the specified and desired type of emoji.